\documentclass{article}

\usepackage[final]{tackling_climate_workshop_style}
\usepackage[utf8]{inputenc} 
\usepackage[T1]{fontenc}    
\usepackage{hyperref}       
\usepackage{url}            
\usepackage{booktabs}       
\usepackage{amsfonts}       
\usepackage{nicefrac}       
\usepackage{microtype}      
\usepackage{amsmath}
\usepackage{xcolor}
\usepackage{multirow}
\usepackage{enumitem}  
\usepackage{calc}
\usepackage{graphicx}
\usepackage{appendix}
\usepackage{mhchem}

\title{Neural Network–enabled Domain-consistent Robust Optimisation for Global \ce{CO2} Reduction Potential of Gas Power Plants}

\author{
\begin{minipage}[t]{\textwidth}
\centering
\begin{tabular}{ccc}
\textbf{Waqar Muhammad Ashraf}$^*$ & \textbf{Talha Ansar} & \textbf{Abdulelah S. Alshehri} \\
University College London & University of Engineering  & King Saud University\\
London, UK & and Technology & Riyadh, Saudi Arabia \\
The Alan Turing Institute &  Lahore, Pakistan \\
London, UK \\
\texttt{ waqar.ashraf.21@ucl.ac.uk } & \texttt{} & \texttt{} \\
\end{tabular}
\end{minipage}
\vspace{1.5em}
\\ 
\begin{minipage}[t]{\textwidth}
\centering
\begin{tabular}{ccc}
\textbf{Peipei Chen} & \textbf{Ramit Debnath} & \textbf{Vivek Dua} \\
\textbf{University of Cambridge} & \textbf{University of Cambridge} & \textbf{University College London} \\
\textbf{Cambridge, UK} & \textbf{Cambridge, UK} & \textbf{London, UK} \\
\texttt{} & \texttt{} & \texttt{} \\
\end{tabular}
\end{minipage}
}
\begin{document}
\maketitle

\begin{abstract}
We introduce a neural network-driven robust optimisation framework that integrates data-driven domain as a constraint into the nonlinear programming technique, addressing the overlooked issue of domain-inconsistent solutions arising from the interaction of parametrised neural network models with optimisation solvers. Applied to a 1180 MW capacity combined cycle gas power plant, our framework delivers domain-consistent robust optimal solutions that achieve a verified 0.76 percentage point mean improvement in energy efficiency. For the first time, scaling this efficiency gain to the global fleet of gas power plants, we estimate an annual 26 Mt reduction potential in \ce{CO2} (with 10.6 Mt in Asia, 9.0 Mt in the Americas, and 4.5 Mt in Europe). These results underscore the synergetic role of machine learning in delivering near-term, scalable decarbonisation pathways for global climate action.  
 
\end{abstract}

\section{{Introduction}}

The International Energy Agency (IEA) identifies efficiency as the “first fuel” of the clean energy transition, capable of delivering over one-third of \ce{CO2} reductions needed by 2030 under its Net Zero Emissions Scenario, while reducing global gas demand by around 50\% \cite{ieaeff}. Enhancing the energy efficiency of gas-fired power plants constitutes a cost-effective and expeditious approach to advancing both climate objectives and energy security within the framework of net-zero transitions \cite{nong2025early,colmenar2018technological}. The energy efficiency gains not only cut emissions but also enhance resilience to fuel supply disruptions, complementing longer-term solutions such as hydrogen co-firing and carbon capture, utilisation and storage \cite{ieanetzero, lu2025quantifying}. In this regard, data-driven modelling and optimisation techniques hold a central role for robust optimisation of gas plants' operation that can yield potential energy efficiency gains \cite{jia2022gas}. However, pure data-centric optimisation analytics can be domain-inconsistent with the plant's operation \cite{chai2025cross,afroze2023domain}, which may undermine the potential energy efficiency gains and, in turn, lower \ce{CO2} discharge from gas power plants.  

This paper presents the following contributions: (i) we formulate a domain-consistent optimisation model that constrains optimal solutions to a data-derived Mahalanobis trust region over operating variables, (ii) we train multi-level artificial neural network (ANN) surrogates (turbines and plant level) and embed them in a nonlinear robust optimisation framework along with process constraints, (iii) we verify the optimal solutions against the plant data from a 1180 MW combined cycle gas power plant (CCGPP) unit and achieve the energy efficiency gain, and (iv) we scale up the gain in plant-level energy efficiency to the global fleet of gas power plants and estimate their annual \ce{CO2} reduction potential, identifying a new decision-led pathway for emissions reduction to support climate action.     

 \subsection{Prior work}
 
 A significant proportion of literature on the application of ML for CCGPP is based on this open-source data set \cite{combined_cycle_power_plant_294}, frequently used to benchmark the performance of proposed ML algorithms \cite{zhang2024prediction, xezonakis2024modelling}. However, the data set omits key operating variables and performance parameters (e.g., thermal efficiency and turbine heat rate), limiting the implementation of ML-driven findings in industrial settings. Recently, some studies have deployed real operational data of gas power plants to carry out artificial neural network (ANN)-based optimisation for the performance parameters by deterministic and evolutionary techniques \cite{ntantis2024optimization, turja2024machine, hai2022proposal}. Yet, the literature studies have not examined the critical interaction of parametrised ANN model with the greedy optimisation solvers, which can produce domain-inconsistent and operationally infeasible solutions. This overlooked issue is the major barrier to adopting ML in the safety-critical industrial environment, where uptake remains historically slow. Beyond methodological advancement, the broader question of ML-enabled \ce{CO2} mitigation from the global fleet of gas power plants remains unexplored, despite its importance for navigating new routes to achieve net-zero and to contribute to climate action.
 
\section{Method}

ANN can approximate nonlinear function space with reasonable accuracy and memory requirement \cite{rumelhart1985learning,rumelhart1986}, and is used for power systems' applications \cite{xezonakis2024modelling,arferiandi2021heat}. In this paper, ANN models are trained to predict performance parameters at the subsystems (Gas Turbine (GT)-1,-2, and Steam Turbine (ST)), and plant-level (CCGPP) operation of a 1180 MW capacity CCGPP. The operating variables at all operating levels of CCGPP are selected based on literature review \cite{polyzakis2008optimum,gulen2019gas} and feedback from the performance engineers. A brief description of the operation of CCGPP is provided in Section \ref{plant_operation}. A data set comprising 577 observations, averaged over 15 minutes each, is collected from the CCGPP. It covers a wide operating range of operating variables and performance parameters (thermal efficiency (TE-\%), turbine heat rate (THR-kJ/kWh) and generated power (Power-MW)), and characterises the operation of CCGPP. The statistics of the data set are provided in Table \ref{tab:abbreviations} and Table \ref{tab:summary_stats}.    

Neural network-driven domain-constrained robust optimisation framework consists of two stages. In the first stage, the trained ANN models are embedded in the optimisation problem to optimise performance parameters, i.e., maximise TE and minimise THR at the set value of Power for CCGPP. The plant-level optimisation problem formulated by nonlinear programming is given as:

\begin{equation}
\begin{aligned}
& Objective \quad function: \min_{\mathbf{x}} f(\mathbf{x}) = - f_{\text{TE}}(\mathbf{x}) + f_{\text{THR}}(\mathbf{x}) \\
& \text{s.t.} \quad h(\mathbf{x}) = 0, \\
& \quad (f_{\text{Power}}(\mathbf{x}) - Power_{\text{Set Point}})^2 < \epsilon, \\
& \quad  f_{\text{Power}}(\mathbf{x}) - \sum_{i=1}^3\mathbf{x_i} < \Delta, \quad (\mathbf{x}-\mu)^\top\Sigma^{-1}(\mathbf{x}-\mu) < \tau^2,  \\
& \quad \mathbf{x} \in \mathbf{X} \subseteq \mathbb{R}^m, \quad  \mathbf{x} = \{x_1, x_2, \dots, x_m\}, \quad \mathbf{x}^L \leq \mathbf{x} \leq \mathbf{x}^U
\end{aligned}
\label{eq:optimization_model}
\end{equation}

here, $h(x)$ is an equality constraint, and represents a trained ANN models. Equal weight is applied to the optimisation terms in the objective function, as they are treated equally for the performance evaluation of thermal power plants \cite{eia2015}. An inequality constraint minimises the squared deviation between the set value of Power ($Power_{\text{ Set Point}}$) and the model-based simulated Power ($f_{\text{Power}}\mathbf{(x)}$) up to $\epsilon$. $\mathbf{x_i}$ is a set of variables (Power-MW) from subsystem level (GT-1,GT-2, ST) and the deviation of their summation from $f_{\text{Power}}(\mathbf{x})$ is kept below $\Delta$-auxiliary power consumption in CCGPP. Whereas, the Mahalanobis distance-based constraint ($\mathbf{x}-\mu)^\top\Sigma^{-1}(\mathbf{x}-\mu)\le \tau^2$) introduces the data-driven domain knowledge of the plant's operation into the optimisation problem for estimating domain-consistent optimal solutions. $\mathbf{x}$ is a set of operating variables and has lower ($\mathbf{x}^L$) and upper ($\mathbf{x}^U$) bounds. With the top-down optimisation approach, the operating conditions of the operating variables at the subsystem level are simulated (see Section \ref{oPTIMISATION} for more details).

In the second stage, the robustness of the estimated solutions at all operating levels of CCGPP is examined by the input perturbation technique \cite{mirjalili2016obstacles}. Gaussian noise on the noise level from 1\% to 5\% of the ranges of operating variables is generated and added with the optimal solutions to construct 10,000 simulated experiments \cite{mirjalili2016obstacles}. The variance threshold is kept below 0.01 to account for robust optimal solutions \cite{mirjalili2015novel} (refer to Section \ref{rob_eva} for more details). Later, the optimal solution(s) are verified on the operation of CCGPP, and the potential gain in energy efficiency is converted into an annual reduction in \ce{CO2} emissions from the CCGPP (refer to Section \ref{Emissions}). Later, ML-led cumulative \ce{CO2} reduction potential from the global fleet of gas power plants is estimated. 

\section{Results}

\subsection{Plant-level energy efficiency gain}

ANN models corresponding to the operating levels of subsystems (GT-1, GT-2, ST) and plant-level (CCGPP) are trained with a coefficient of determination ($R^2$) equal to or greater than 0.85 in the test data set. Details about the predictive performance of the models are provided in Table \ref{Tab_model_train}. The plant-level optimisation problem is solved on two set-point values of Power, i.e., ($Power_{\text{ Set Point}} = 950 \space  MW, 1090 \space MW$) by Interior Point solver in Pyomo \cite{floudas1995nonlinear}. The two operating levels of Power are chosen because (i) gas power plants operate nearly at full design capacity \cite{gonzalez2018review}, and (ii) to demonstrate the efficacy of the ANN-based optimisation framework to adapt to variable power demand.

\begin{figure}[htp]
    \centering
    \includegraphics[width=0.6\linewidth]{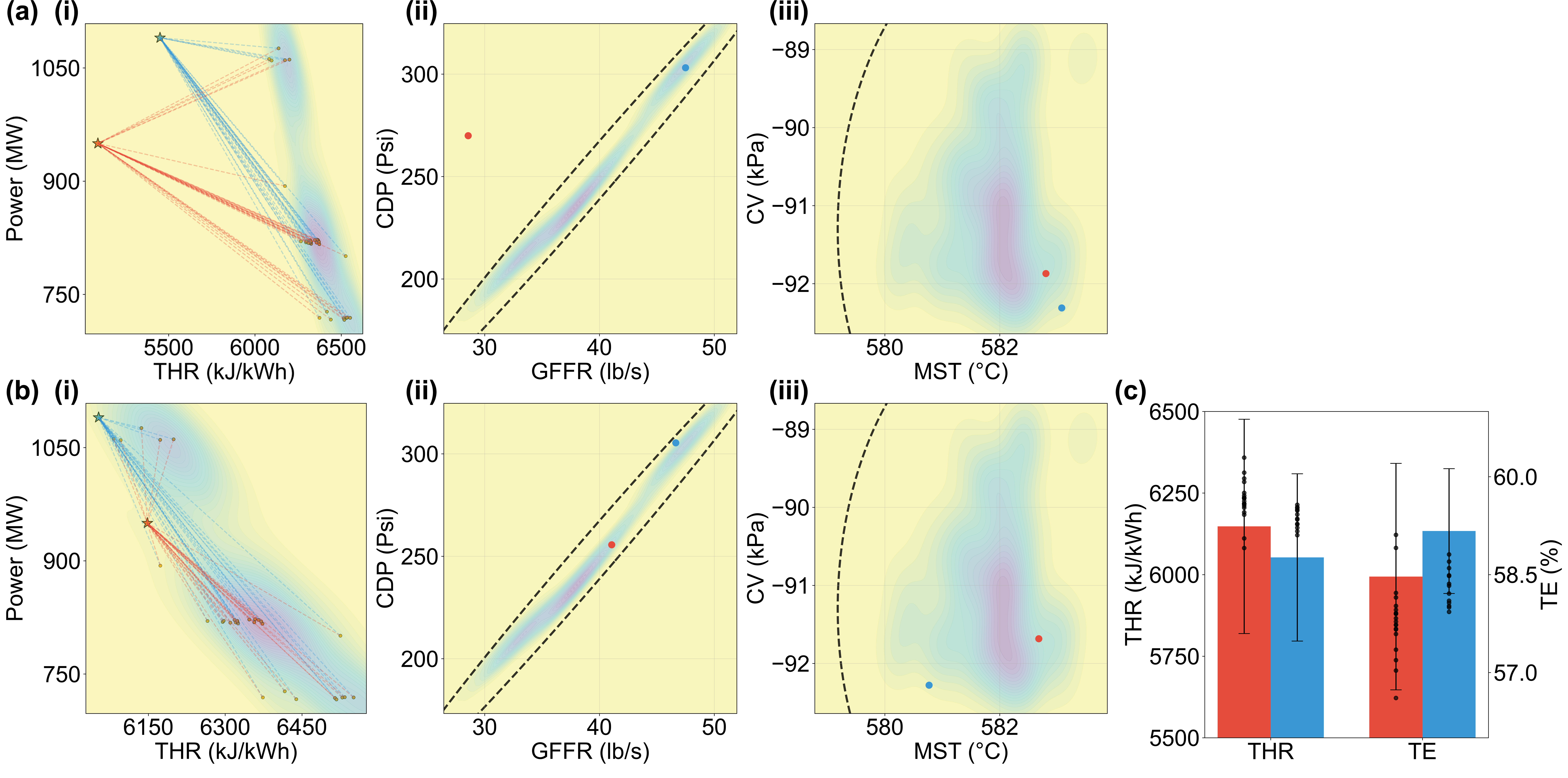}
    \caption{Multi-level optimisation of CCGPP. Solver convergence for optimisation problem solved (a)(i) without and (b)(i) with Mahalanobis constraint for $Power_{\text{ Set Point}}$ of 950 MW (red) and 1090 MW (blue). (a)(ii-iii) and (b)(ii-iii) Mapping the optimal solutions, and (c) comparing the optimal solutions with the actual data of power plant.  }
    \label{Fig:opt}
\end{figure}

The optimisation problem \ref{eq:optimization_model} is solved with respect to different initial guesses and without embedding Mahalanobis distance-based constraint to observe the quality of the optimal solution in terms of optimising the performance parameters and domain consistency. It is noted that the optimisation solver estimates the optimal values of TE and THR (refer to Figure \ref{Fig:opt}(a)(i)) that are beyond their nominal operating ranges, as mentioned in Table \ref{tab:summary_stats}. Domain-inconsistent optimum solution is estimated (shown in Figure \ref{Fig:opt}(a)(ii) for set value of Power of 950 MW (red colour)) as it is mapped significantly outside the operating envelope of the two correlated features (GFFR \& CDP of GT-1). Feasible and domain-consistent optimal solutions are obtained after solving the Mahalanobis distance-based constrained optimisation problem \ref{eq:optimization_model} (refer to Figure \ref{Fig:opt}(b)(i)-(iii). Out of the feasible solutions, the optimal solution is selected that exhibits the lowest variance, even below the threshold (0.01), in the plant-level performance parameters (refer to Section \ref{rob_eva} to get specific details).

The effectiveness of the estimated solution is verified on the actual data of CCGPP. TE \& THR at the plant level for the power value of 950 MW and 1090 MW are achieved with the mean absolute percentage error of 0.1\% \& 0.04\%, and 0.04\% \& 0.5\%, respectively. Comparing the improvement in TE with the plant's operation corresponding to power generation of 950 MW and 1090 MW (refer to Figure \ref{Fig:opt}(d)), we identify a mean improvement in TE of 0.76 percentage point (pp), achieved through ML-led synchronised operation optimisation of CCGPP.

\subsection{{Emissions reduction from global fleet of CCGPP}}

\begin{figure}[htp]
    \centering
    \includegraphics[width=0.6\linewidth]{ 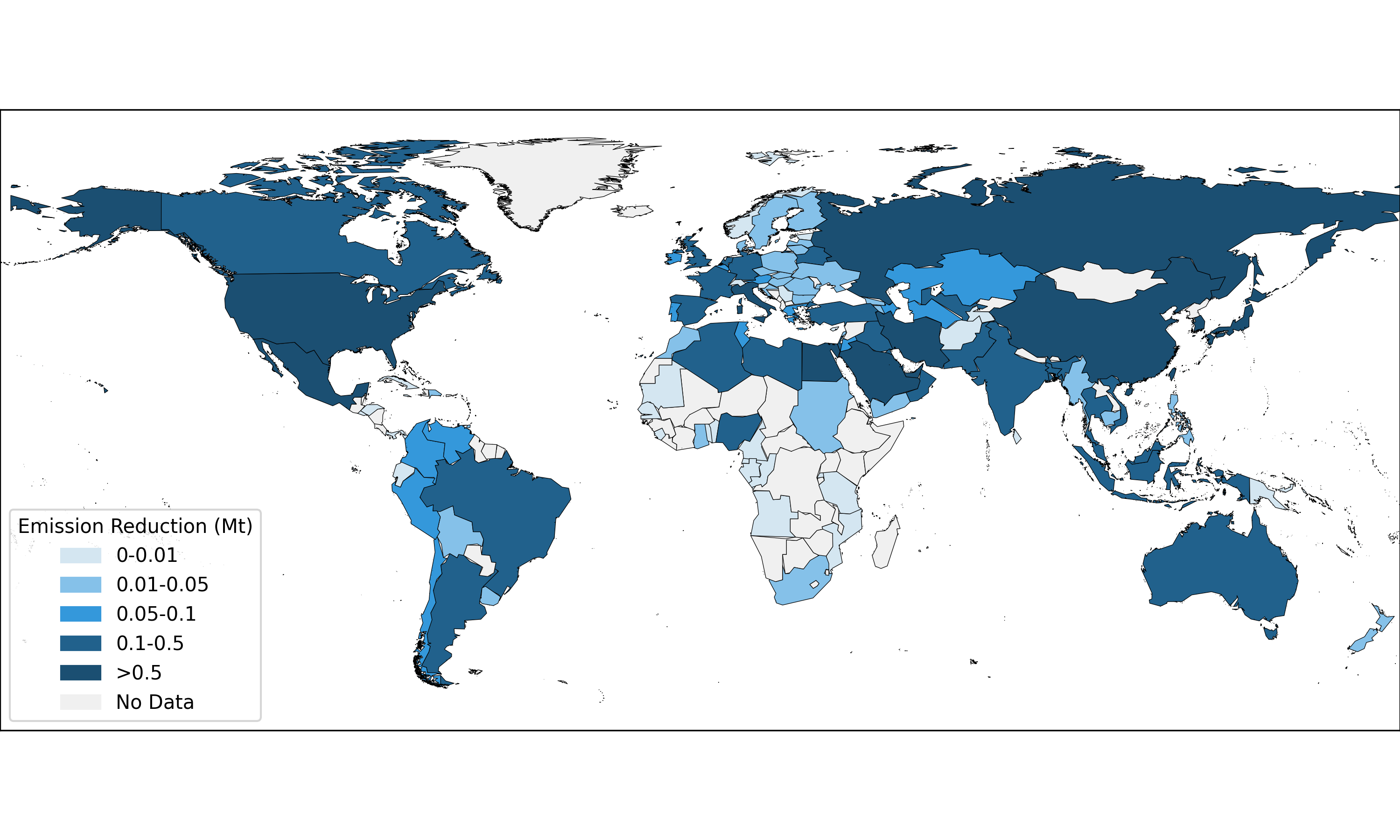}
    \caption{Global \ce{CO2} reduction potential from gas power plants. }
    \label{Fig:ger}
\end{figure}

Building on ANN-led estimated efficiency improvement potential of the CCGPP unit, we consider 0.76 $\pm$ 0.5 pp (minimum and maximum energy efficiency gain possible in the global fleet of CCGPP \cite{Globalgas} due to heterogeneous factors, including combustion technology, fuel quality, capacity of plant etc.,) and extend this range to estimate associated global reduction in carbon emissions. The results suggest that targeted efficiency upgrades could collectively avoid around 26 Mt \ce{CO2} emissions annually (Figure \ref{Fig:ger}), a reduction comparable to taking several million cars off the road each year. The largest national potential lies in the United States (7.1 Mt), where a combination of high installed capacity and relatively mature infrastructure could deliver over a quarter of the global total. Significant opportunities also exist in China (1.7 Mt), Russia (1.5 Mt), and Japan (1.1 Mt), where efficiency gains could meaningfully offset growing electricity demand and reduce reliance on imported fuels.

When viewed at a regional scale, Asia emerges as the leading contributor to potential savings (10.6 Mt), driven primarily by Eastern Asia (3.7 Mt), where China’s extensive gas power fleet dominates the picture. The Americas follow closely (9.0 Mt), with most of the reductions concentrated in North America (7.5 Mt) due to its large, high-utilisation plants. Europe also offers notable opportunities (4.5 Mt), particularly in Eastern Europe (1.8 Mt) where older generation units still prevail. In contrast, Africa’s contribution is more modest (1.5 Mt), reflecting both its smaller installed base and lower operating hours. These patterns highlight that efficiency improvements in gas-fired generation not only cut emissions in the near term but also enhance fuel efficiency, reduce operating costs, and strengthen energy resilience, providing a practical bridge between today’s power systems and the deeper decarbonisation required for net-zero pathways.

\section*{Conclusion}

This research addresses the domain-inconsistency issue arising from the interaction of parametrised neural network models with goal-oriented optimisation solvers to estimate robust optimal solutions. We demonstrate the adaptability of the developed neural network-driven domain-consistent robust optimisation framework for operation optimisation of a 1180 MW capacity CCGPP and achieve a 0.76 percentage point mean efficiency gain for the power plant's operation. Scaling the efficiency gain to the global fleet of gas power plants, an annual reduction potential of 26 Mt \ce{CO2} is estimated, with the largest potential in Asia (10.6 Mt), the Americas (9.0 Mt) and Europe (4.5 Mt).

These findings suggest that machine learning–enabled robust optimisation not only enhances the resilience of plant performance but also offers scalable short-term decarbonisation pathway, complementing long-term solutions and expanding the role of artificial intelligence in climate action.
\section*{Acknowledgments}

Waqar Muhammad Ashraf acknowledges that this work was supported by The Alan Turing Institute’s Enrichment scheme. Waqar Muhammad Ashraf also acknowledges the funding received from The Punjab Education Endowment Fund (CMMS-PhD-2021-006) for his PhD at University College London, UK. The mentorship received from Arun K Choudhary as part of The Climate Change AI mentorship program is also acknowledged. 

\bibliographystyle{unsrt}  
\bibliography{reference}

\begin{appendices}
\clearpage
\section*{Appendix}
\numberwithin{equation}{section}
\numberwithin{figure}{section}
\renewcommand{\thefigure}{\thesection\arabic{figure}}
\renewcommand{\thetable}{\thesection\arabic{table}}
\renewcommand{\theequation}{\thesection\arabic{equation}}
\section{Brief description of combined cycle gas power plant} 
\label{plant_operation}
The 1180 MW capacity combined cycle gas power plant (CCGPP) comprises two gas turbines (395 MW capacity each) and the exhaust gases from the gas turbines operate a steam turbine system (388 MW capacity). Turbine heat rate (THR - kJ/kWh), thermal efficiency (TE - \%),  and Power (Power - MW) are measured at the output of gas turbines while Power and THR are measured at the output of steam turbine system. At the plant-level, TE, THR and Power are measured to evaluate the performance of CCGPP.

 At the gas turbine (GT) level, the selected operating features are as follows: temperature of air at the discharge of compressor (CDT-$^{\circ}$F), pressure of air at the discharge of compressor (CDP-Psi), Flow Rate of Gas (GFFR-lb/s), Performance Heater Gas Outlet Temperature (PHGOT-$^{\circ}$F), Temperature of gas at entrance of combustion chamber (FGT-$^{\circ}$F), Atmospheric Temperature (AT-$^{\circ}$C), Atmospheric Pressure (AP-hPa) and Atmospheric Humidity (AH-\%). The same operating features are used for two gas turbines. Whereas, the operating features selected for steam turbine system are as follows: Flue Gas Temperature at the inlet of Heat Recovery Steam Generator (FGT\_HRSG-$^{\circ}$C), Feed Water Temperature at Intermediate Pressure (IP) Economizer (FWT-$^{\circ}$C), Reheat Steam Pressure at IP inlet (RHP-MPa), Reheat Steam Temperature at IP inlet (RST-$^{\circ}$C), Reheat Steam Flow Rate at IP inlet (RSF-t/h), Steam Flow Rate at the inlet of Low Pressure (LP) Turbine (LPST-t/h), Steam Pressure at the inlet of LP Turbine (LPSP-MPa), Steam Temperature at the inlet of LP Turbine (LPST-$^{\circ}$C), and Condenser Vacuum (CV-kPa). The outputs of gas turbines (TE, THR and Power) and steam turbine (THR and Power) are deployed to model the plan-level performance parameters (TE, THR and Power) of CCGPP.
 
 After the selection of the operating variables and the corresponding performance parameters at the sub-system level (GT-1, GT-2, ST) and plant-level (CCGPP), a dataset associated with the operating features is collected from the power plant. The data set covers a wide operating range of the power plant's operation. The descriptive statistics of the collected data are provided in Table \ref{tab:summary_stats}. 

\renewcommand{\thetable}{A\arabic{table}}
\setcounter{table}{0} 
\begin{table}[ht]
\centering
\caption{Nomenclature}
\label{tab:abbreviations}
\begin{tabular}{ll}
\toprule
\textbf{Abbreviation} & \textbf{} \\
\midrule
AH      & Atmospheric Humidity (\%) \\
ANN      & Artificial Neural Network \\
AP      & Atmospheric Pressure (hPa) \\
AT      & Atmospheric Temperature (°C) \\
CCGPP   & Combined Cycle Gas Power Plant \\
CDP     & Pressure of air at the discharge of compressor (Psi) \\
CDT     & Temperature of air at the discharge of compressor (°F) \\
CV      & Condenser Vacuum (kPa) \\
FGT     & Temperature of gas at entrance of combustion chamber (°F) \\
FGT\_HRSG & Flue Gas Temperature at the inlet of HRSG\\
FWT     & Feed Water Temperature at IP Economizer (°C) \\
GFFR    & Flow Rate of Gas (lb/s) \\
HP      & High Pressure \\
HRSG    & Heat Recovery Steam Generator\\
IP      & Intermediate Pressure \\
LP      & Low Pressure \\
LPSF    & Steam Flow Rate at the inlet of LP Turbine (t/h) \\
LPSP    & Steam Pressure at the inlet of LP Turbine (MPa) \\
LPST    & Steam Temperature at the inlet of LP Turbine (°C) \\
ML      & Machine Learning \\
MSFR    & Main Steam Flow Rate at HP Inlet (t/h) \\
MSP     & Main Steam Pressure at HP Inlet (MPa) \\
MST     & Main Steam Temperature (°C) \\
PHGOT   & Performance Heater Gas Outlet Temperature (°F) \\
Power   & Power (MW) \\
PP      & Percentage Point \\
RSF     & Reheat Steam Flow Rate at IP Inlet (t/h) \\
RSP     & Reheat Steam Pressure at IP Inlet (MPa) \\
RST     & Reheat Steam Temperature at IP Inlet (°C) \\
TE      & Thermal Efficiency (\%) \\
THR     & Heat Rate (kJ/kWh) \\
\bottomrule
\end{tabular}
\end{table}

\begin{table}[ht]
\caption{Summary statistics for variables across different systems with units}
\label{tab:summary_stats}
\centering
\begin{tabular}{llllrrrr}
\toprule
System & Variable & Unit & Minimum & Mean & Maximum & Standard Deviation \\
\midrule
\multirow{11}{*}{\textbf{GT-1}} 
 & CDT & °F & 813 & 861 & 926 & 34.3 \\
 & CDP & Psi & 186 & 248 & 312 & 36.9 \\
 & GFFR & lb/s & 29 & 39 & 50 & 5.7 \\
 & PHGOT & °F & 400 & 411 & 425 & 2.5 \\
 & FGT & °F & 484 & 513 & 535 & 15.0 \\
 & AT & °C & 20 & 26 & 34 & 3.7 \\
 & AP & hPa & 983 & 988 & 992 & 2.0 \\
 & AH & \% & 34 & 66 & 98 & 14.2 \\
 \cmidrule(lr){2-7}
 & Power & MW & 186 & 297 & 396 & 59.4 \\
 & THR & kJ/kWh & 8377 & 9267 & 11022 & 580.0 \\
 & TE & \% & 32.7 & 39.0 & 43.0 & 2.4 \\
\midrule
\multirow{11}{*}{\textbf{GT-2}} 
 & CDT & °F & 829 & 872 & 927 & 34.4 \\
 & CDP & Psi & 202 & 246 & 311 & 37.2 \\
 & GFFR & lb/s & 31 & 39 & 48 & 5.5 \\
 & PHGOT & °F & 404 & 411 & 418 & 1.5 \\
 & FGT & °F & 494 & 523 & 541 & 11.5 \\
 & AT & °C & 20 & 26 & 34 & 3.7 \\
 & AP & hPa & 983 & 988 & 992 & 2.0 \\
 & AH & \% & 34 & 66 & 98 & 14.2 \\
 \cmidrule(lr){2-7}
 & Power & MW & 216 & 292 & 390 & 56.6 \\
 & THR & kJ/kWh & 8471 & 9308 & 10311 & 506.0 \\
 & TE & \% & 35.0 & 38.8 & 42.5 & 2.1 \\
\midrule
\multirow{14}{*}{\textbf{ST}}
 & FGT & °C & 629 & 659 & 673 & 15.9 \\
 & FWT & °C & 153 & 159 & 167 & 4.4 \\
 & MSP & MPa & 11.5 & 13.1 & 15.2 & 1.2 \\
 & MSFR & t/h & 483 & 623 & 731 & 64.5 \\
 & MST & °C & 579 & 582 & 583 & 0.7 \\
 & RSP & MPa & 2.56 & 2.88 & 3.26 & 0.2 \\
 & RST & °C & 580 & 582 & 584 & 0.7 \\
 & RSF & t/h & 529 & 676 & 797 & 71.3 \\
 & LPSF & t/h & 596 & 758 & 906 & 85.7 \\
 & LPSP & MPa & 0.37 & 0.45 & 0.55 & 0.1 \\
 & LPST & °C & 314 & 319 & 324 & 2.5 \\
 & CV & kPa & -92.3 & -91.0 & -89.0 & 0.9 \\
 \cmidrule(lr){2-7}
 & Power & MW & 260 & 297 & 344 & 25.0 \\
 & THR & kJ/kWh & 2642 & 3048 & 3375 & 110.5 \\
\midrule
\multirow{3}{*}{\textbf{CCGPP}}
 & Power & MW & 680 & 859 & 1094 & 130.9 \\
 & THR & kJ/kWh & 6052 & 6363 & 6723 & 149.9 \\
 & TE & \% & 53.6 & 56.6 & 59.5 & 1.3 \\
\bottomrule
\end{tabular}
\end{table}

\section{Training of ANN models for CCGPP} \label{model_training}
ANN models are trained under hyperparameter tuning. We train shallow three-layered ANN models since they can model any nonlinear function space as long as reasonable number of neurons are embedded in the hidden layer \cite{bishop2006pattern}. Sigmoid Linear Unit (SiLU) based activation function is implemented on the hidden layer of ANN for its smooth and non-monotonicity properties compared with the rectified linear unit \cite{elfwing2018sigmoid,hendrycks2016gaussian}. Learning rate, neurons number in the hidden layer, and $L_1$ regularisation parameter ($\lambda_1$) are optimised by Tree-structured Parzen Estimator solver that adopts Bayesian optimisation approach \cite{bergstra2015hyperopt}. Later, the optimal values of hyperparameters are embedded in the architecture of ANN and the parameters of ANN models are tuned by Adaptive Moment Estimation \cite{kingma2014adam}. The data partition is made on the ratio of 0.2 and 0.8 for testing and training datasets, respectively, for the evaluation of model's predictions. The predictive performance of the trained ANN models on the training and testing datasets is evaluated by co-efficient of determination ($R^2$) and root mean square error ($RMSE$). The two metrics are commonly used to evaluate the modelling performance of ML models \cite{yuan2021applied, akdacs2023data}.  The predictive performance of the trained models is provided in Table \ref{tab:ANN_metrics}.  

\renewcommand{\thetable}{B\arabic{table}}
\setcounter{table}{0}
\begin{table}[ht]
  \caption{Performance metrics of trained ANN models for multi-level operation of gas power plant.}
  \label{tab:ANN_metrics}
  \centering
  \begin{tabular}{llcccccccc}
    \toprule
    & & \multicolumn{2}{c}{GT-1} & \multicolumn{2}{c}{GT-2} & \multicolumn{2}{c}{ST} & \multicolumn{2}{c}{CCGPP} \\
    \cmidrule(lr){3-4} \cmidrule(lr){5-6} \cmidrule(lr){7-8} \cmidrule(lr){9-10}
    & & R$^2$ & RMSE & R$^2$ & RMSE & R$^2$ & RMSE & R$^2$ & RMSE \\
    \midrule
    \multirow{2}{*}{\textbf{Power (MW)}} 
        & Train & 0.99 & 0.85 & 0.99 & 0.83 & 0.99 & 1.25 & 0.99 & 7.94 \\
        & Test  & 0.99 & 0.93 & 0.99 & 0.81 & 0.99 & 1.25 & 0.99 & 7.92 \\
    \multirow{2}{*}{\textbf{THR (kJ/kWh)}} 
        & Train & 0.88 & 203.31 & 0.89 & 163.46 & 0.95 & 23.43 & 0.89 & 49.27 \\
        & Test  & 0.84 & 221.70 & 0.86 & 181.03 & 0.95 & 24.44 & 0.85 & 54.46 \\
    \multirow{2}{*}{\textbf{TE (\%)}} 
        & Train & 0.97 & 0.38 & 0.97 & 0.35 & -- & -- & 0.94 & 0.30 \\
        & Test  & 0.96 & 0.40 & 0.96 & 0.37 & -- & -- & 0.94 & 0.29 \\
    \bottomrule
    \label{Tab_model_train}
  \end{tabular}
\end{table}

\section{Optimisation problem for subsystems of CCGPP} \label{oPTIMISATION}
A nonlinear programming framework is implemented to simulate the optimal values of operating variables for the performance parameters related to the operating level of subsystems (GT-1, GT-2, ST) of CCGPP. The outputs of subsystems are deployed to predict the plant-level performance parameters. Under the top-down approach, we first optimise TE and THR (plant-level performance parameters of CCGPP) on the set values of Power, and estimate the optimal input conditions which are actually the outputs of subsystems (GT-1, GT-2 and ST). In order to achieve the optimal output values of the performance parameters of subsystems, we formulate an optimisation problem which is defined similar to Equation \ref{eq:optimization_model}. However, the objective function along with process constraints for GT and ST is defined differently and is written as follows:

\begin{equation}
\begin{aligned}
\text{For GT: }
& \min_{\mathbf{x}} \quad f(\mathbf{x}) =  (f_{\text{Power}}(\mathbf{x}) - Power_{\text{Set Point}})^2  \\
& \text{s.t.} \quad h(\mathbf{x}) = 0, \\
& \quad (f_{\text{TE}}(\mathbf{x}) - TE_{\text{Set Point}})^2 < \epsilon, \\
& \quad (f_{\text{THR}}(\mathbf{x}) - THR_{\text{Set Point}})^2 < \epsilon, \\
& \quad (\mathbf{x}-\mu)^\top\Sigma^{-1}(\mathbf{x}-\mu) < \tau^2, \\
& \quad \mathbf{x} = \{x_1, x_2, \dots, x_p\}, \quad \mathbf{x} \in \mathbf{X} \subseteq \mathbb{R}^p,  \quad \mathbf{x}^L \leq \mathbf{x} \leq \mathbf{x}^U
\end{aligned}
\label{eq:GT-OPT_MODEL}
\end{equation}

\begin{equation}
\begin{aligned}
\text{For ST: }
& \min_{\mathbf{x}} \quad f(\mathbf{x}) =  (f_{\text{Power}}(\mathbf{x}) - Power_{\text{Set Point}})^2 + (f_{\text{THR}}(\mathbf{x}) - THR_{\text{Set Point}})^2 \\
& \text{s.t.} \quad h(\mathbf{x}) = 0, \\
& \quad (\mathbf{x}-\mu)^\top\Sigma^{-1}(\mathbf{x}-\mu) < \tau^2, \\
& \quad \mathbf{x} = \{x_1, x_2, \dots, x_q\}, \quad \mathbf{x} \in \mathbf{X} \subseteq \mathbb{R}^q, \quad \mathbf{x}^L \leq \mathbf{x} \leq \mathbf{x}^U
\end{aligned}
\label{eq:ST-OPT_MODEL}
\end{equation}

here, $\mu$ is mean vector while $\Sigma$ is the covariance matrix computed on data associated with $\mathbf{x}$. $\tau$ controls the feasible space around the mean-centroid of the joint-distribution of $\mathbf{x}$ where optimisation solver searches for optimal solution. The number of elements in $\mathbf{x}$ are defined as per the operating variables related with operating-level of power plant (GT-1, GT-2, ST, CCGPP).

The defined optimisation problems for GT-1, GT-2 and ST are solved for different values of $\epsilon$ and $\tau$ such that the optimal values of their performance parameters are closely matched with those of plant-level optimal input conditions (same as performance parameters of subsystems). This allows the operation of the subsystems at the optimal process conditions such that their performance parameters closely predict the performance parameters at the plant-level operation of CCGPP. Mahalanobis constraint guides the optimisation solver for estimating the domain-consistent solutions and enhances their implementability in the operation of the power plant.

\section{Robustness evaluation of optimal solutions} \label{rob_eva}
The following equations are used to evaluate the robustness of the optimal solutions ($x^*$) obtained for sub-system and plant-level performance parameters. 

\begin{align}
& F(x^*) =  \frac{\sum_{k=1}^Hf(x^*+\delta_k)}{H} \\
& V(x^*) =  \frac{||F(x^*) - f(x^*)||}{||f(x^*)||} < \epsilon
\label{eq:robustness}
\end{align}

here, $\delta$ refers to the noise observations produced on the noise level (varied from 1\% to 5\% of the operating ranges of the variables). $k$ indicates the number of noise observations and are set to 10000, i.e., $H$. The generated Gaussian noise observations ($\delta_k$) are added with the optimal solution ($x^*$) to investigate the variation in the function space of the performance parameter driven by perturbing $x^*$ \cite{mirjalili2015novel}. The vicinity of $x^*$ is comprehensively explored through the constructed experiments ($x^*+\delta_k$) and the mean response of the function ($F(x^*)$) is computed. Later, variance produced due to constructed experiments ($V(x^*)$) is computed. The estimated solution can be regarded as robust if $V(x^*)$ is less than $\varepsilon$. 

The optimal solution estimated for Power of 950 MW remains robust for 1\% noise level of operating ranges for all operating levels of CCGPP. Whereas, the optimal solution absorbs 5\% noise level for the set value of power of 1090 and still remains robust for the operation of CCGPP. For Power of 950 MW and 1090 MW, the variance computed in the plant-level performance parameters, i.e., Power, TE and THR is as follows: 0.0002, 0.0002, 0.0001 and 0.0009, 0.0062, 0.0004 respectively. 

\section{Emission reduction} \label{Emissions}

The gain in energy efficiency is converted into annual reduction in \ce{CO2} emissions from the power plant. Capacity factor, power plant's generation capacity,  and operating hours of the power plant are important factors in determining the emissions discharge. The annual reduction in \ce{CO2} emissions from the global gas power plants (refer to Equation \ref{eq:emission}) is calculated as follows:

\begin{equation}
\text{CO}_2 \text{ reduction} =
\textit{Capacity} \times \textit{Capacity factor} \times \textit{Hours} \times \textit{Emission factor} \times
\left( 1 - \frac{1}{1 + \textit{efficiency improvement}} \right)
\label{eq:emission}
\end{equation}

\begin{figure}[htp]
    \centering
    \includegraphics[width=1.0\linewidth]{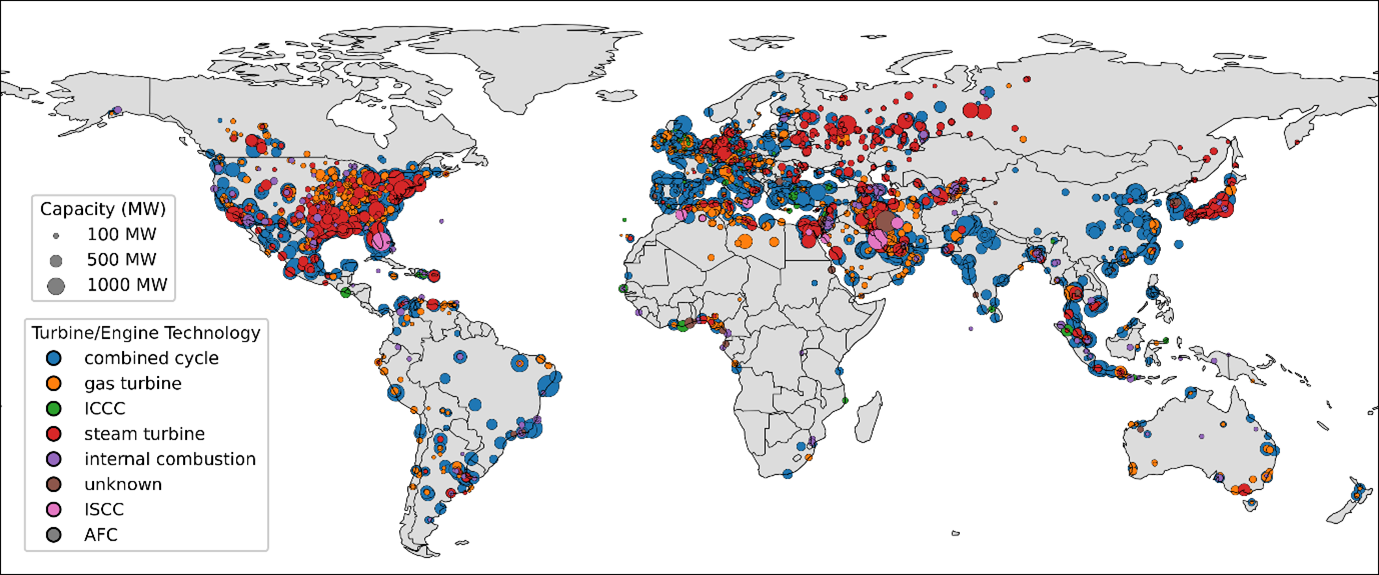}
    \caption{Distribution of global gas power plants.  }
    \label{Fig:ggp}
\end{figure}

The capacity factor and emission factor are taken as 0.5 and 0.4 ton of \ce{CO2} per MWh, respectively. In Figure \ref{Fig:ggp}, the engine technologies are categorised into the following types:
\begin{itemize}
    \item Internal Combustion: Power plants using reciprocating internal combustion engines fueled by natural gas or other fuels. They are typically small to medium scale, flexible, and suitable for distributed generation or peaking applications.
\end{itemize}
\begin{itemize}
    \item Gas Turbine: Plants where natural gas is combusted to drive a gas turbine connected to a generator. They offer fast start-up and are widely used for peak-load or backup power.
\end{itemize}
\begin{itemize}
    \item Steam Turbine (ST): Gas is burned to produce steam, which then drives a steam turbine. These plants are usually less efficient on their own compared to combined-cycle systems.
\end{itemize}
\begin{itemize}
    \item Combined Cycle (CC): A configuration that combines a gas turbine with a steam turbine, using the waste heat from the gas turbine to produce steam. This significantly improves overall efficiency, making CC plants the dominant form of modern gas power generation.
\end{itemize}
\begin{itemize}
    \item ICCC (Internal Combustion Combined Cycle): Internal combustion engines are integrated with a combined cycle system to enhance efficiency and reduce emissions.
\end{itemize}
\begin{itemize}
    \item ISCC (Integrated Solar Combined Cycle): A hybrid design that integrates solar thermal energy with a conventional combined cycle gas turbine plant, improving fuel efficiency and reducing \ce{CO2} emissions.
\end{itemize}
\begin{itemize}
    \item AFC (Allam–Fetvedt Cycle): A novel natural gas power cycle that uses supercritical \ce{CO2} as the working fluid, enabling near-zero emissions with inherent carbon capture and high efficiency.
\end{itemize}

\end{appendices}

\end{document}